\documentclass[letterpaper]{article} 
\usepackage{aaai2026}  
\usepackage{times}  
\usepackage{helvet}  
\usepackage{courier}  
\usepackage[hyphens]{url}  
\usepackage{graphicx} 
\usepackage{natbib}  
\usepackage{caption} 
\usepackage{algorithm}
\usepackage{algorithmic}
\usepackage{amsmath}  
\usepackage{amsfonts} 
\usepackage{booktabs} 
 \usepackage{multirow} 
 \usepackage{makecell} 
\usepackage{newfloat}
\usepackage{listings}
\DeclareCaptionStyle{ruled}{labelfont=normalfont,labelsep=colon,strut=off} 
\floatstyle{ruled}
\newfloat{listing}{tb}{lst}{}
\floatname{listing}{Listing}
\title{Blur-Robust Detection via Feature Restoration: An End-to-End Framework for Prior-Guided Infrared UAV Target Detection}

\author{
    Xiaolin Wang\textsuperscript{\rm 1},
    Houzhang Fang\textsuperscript{\rm 1}\thanks{Corresponding author.},
    Qingshan Li\textsuperscript{\rm 1},
    Lu Wang\textsuperscript{\rm 1},
    Yi Chang\textsuperscript{\rm 2},
    Luxin Yan\textsuperscript{\rm 2}\\
}
\affiliations{
    \textsuperscript{\rm 1}School of Computer Science and Technology, Xidian University, China\\
    \textsuperscript{\rm 2}School of Artificial Intelligence and Automation, Huazhong University of Science and Technology, China\\
{\small wxl@stu.xidian.edu.cn,houzhangfang@xidian.edu.cn,qshli@mail.xidian.edu.cn,wanglu@xidian.edu.cn,\{yichang,yanluxin\}@hust.edu.cn}
}

\begin{document}

\maketitle
\begin{abstract}
Infrared unmanned aerial vehicle (UAV) target images often suffer from motion blur degradation caused by rapid sensor movement, significantly reducing contrast between target and background. Generally, detection performance heavily depends on the discriminative feature representation between target and background. Existing methods typically treat deblurring as a preprocessing step focused on visual quality, while neglecting the enhancement of task-relevant features crucial for detection. Improving feature representation for detection under blur conditions remains challenging. In this paper, we propose a novel \textbf{J}oint \textbf{F}eature-\textbf{D}omain \textbf{D}eblurring and \textbf{D}etection  end-to-end framework, dubbed JFD\textsuperscript{3}. We design a dual-branch architecture with shared weights, where the clear branch guides the blurred branch to enhance discriminative feature representation. Specifically, we first introduce a lightweight feature restoration network, where features from the clear branch serve as feature-level supervision to guide the blurred branch, thereby enhancing its distinctive capability for detection. 
We then propose a frequency structure guidance module that refines the structure prior from the restoration network and integrates it into shallow detection layers to enrich target structural information. Finally, a feature consistency self-supervised loss is imposed between the dual-branch detection backbones, driving the blurred branch to approximate the feature representations of the clear one. We also construct a benchmark, named IRBlurUAV, containing 30,000 simulated and 4,118 real infrared UAV target images with diverse motion blur. Extensive experiments on IRBlurUAV demonstrate that JFD\textsuperscript{3} achieves superior detection performance while maintaining real-time efficiency. 
\end{abstract}

\begin{links}
\link{Code}{https://github.com/IVPLaboratory/JFD3}
\end{links}

\section{Introduction}

\begin{figure}[t]
\centering
\includegraphics[width=0.9\columnwidth]{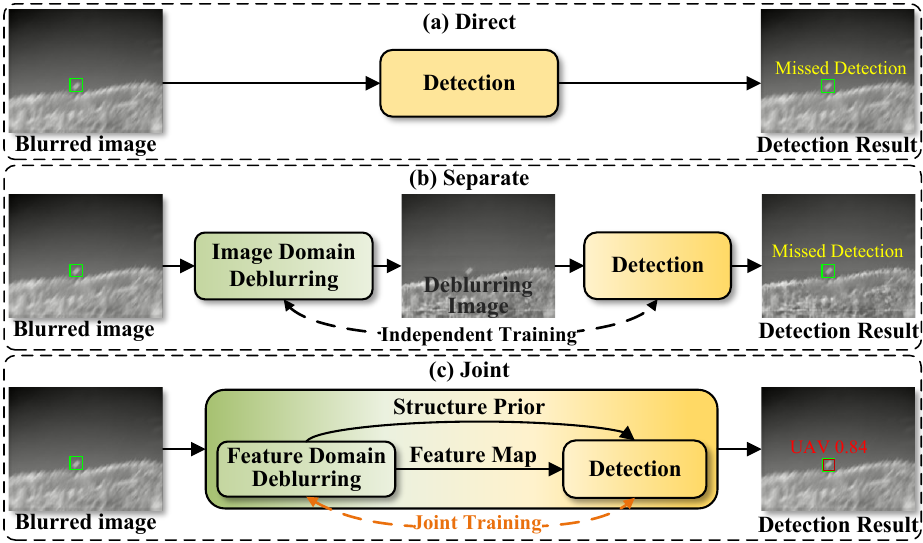} 
\caption{Three strategies for UAV target detection under motion blur. (a) Direct: The detector directly processes blurred images. (b) Separate: Image-domain deblurring serves as a preprocessing step before detection. (c) Joint:  Feature-domain deblurring and detection are simultaneously addressed in an end-to-end framework. Our JFD\textsuperscript{3} jointly handles both tasks and leverages structural priors from the deblurring network to enhance the feature representation of the detection network.}   
\label{fig1}
\end{figure}

Infrared unmanned aerial vehicle (UAV) target (IRUT) detection plays a vital role in many applications, 
such as UAV surveillance and reconnaissance missions, due to its all-day operability and robustness to varying lighting conditions \cite{fang2023differentiated}.
However, motion blur frequently occurs in infrared imagery due to abrupt platform movements initiated to swiftly track fast-moving UAVs or sudden mechanical vibrations (see the left of Figure \ref{fig1}). 
Such motion blur is frequent and often unavoidable in long-term UAV surveillance, posing significant challenges to accurate target detection.
Moreover, IRUT typically exhibit weak features and are embedded in complex clutter backgrounds. Motion blur further diminishes the contrast between targets and surroundings, making discriminative feature extraction more difficult.
In recent years, remarkable progress has been made in both image deblurring \cite{nah2017deep, tao2018scale, kupyn2018deblurgan, kupyn2019deblurgan, lin2020learning} and IRUT detection \cite{fang2022infrared, Fang2024SCINet}. However, most existing approaches treat deblurring and detection as two separate tasks, addressed independently with different objectives. To the best of our knowledge, there has been no prior work that specifically addresses IRUT detection under motion blur conditions.

To address the above problem, a straightforward approach is to directly apply detectors \cite{zhao2024detrs} on blurred images. However, blur degradation significantly reduces the contrast between UAV targets and backgrounds, leading to frequent missed detections (see Figure \ref{fig1}(a)).
Alternatively, deblurring \cite{mao2023DeepRFT} can be applied as a preprocessing step before detection \cite{zhao2024detrs}  (see Figure \ref{fig1}(b)). However, this pipeline suffers from several limitations. First, deep learning-based deblurring techniques are computationally complex, introducing substantial latency and limiting their applicability in time-critical UAV surveillance tasks. 
Second, these methods are typically optimized for visual enhancement rather than task-specific feature restoration, which may introduce imperceptible noise that can negatively impact detection performance \cite{li2023detection, xu2024learning}.
Recently, some studies \cite{li2023detection, xu2024learning,li2025DREB} have explored the joint optimization of low-level and high-level vision tasks. However, most of these efforts focus on adverse weather conditions such as fog and are primarily conducted in the visible spectrum towards general object categories. In contrast, the unique challenges of motion blur in IRUT detection remain largely underexplored.

To bridge the gap between infrared low-level deblurring and high-level IRUT detection, we propose the \textbf{J}oint \textbf{F}eature-\textbf{D}omain \textbf{D}eblurring and \textbf{D}etection Network (JFD\textsuperscript{3}). It adopts a dual-branch architecture during training, where a clear-image branch supervises a blurred-image branch to jointly optimize feature restoration and detection. At inference time, only the blurred branch is retained to enable efficient inference.
Specifically, to address the limitations of conventional image-domain deblurring, which often introduces redundancy and lacks detection-oriented awareness, we design a lightweight feature restoration network guided by a clear feature-domain deblurring  branch and jointly trained with the detection network. This network efficiently enhances degraded representations critical for detection. 
Furthermore, to improve structural perception under motion blur, we design a frequency structure guidance module. It first extracts high-frequency detail features using an adaptive high-pass filtering module, and then refines structural information via a detail-preserving attention mechanism.  The refined structural prior is then injected between the stem and stage 1 of the detection backbone, compensating for missing structural cues in blurred images and improving the discriminability of IRUT.
Finally, to enhance the backbone's ability to extract meaningful target features from degraded inputs, we introduce a feature consistency self-supervised loss between the blurred and clear branches. This constraint  encourages the blurred branch to approximate the clean branch in feature space, enabling more accurate target discrimination under blur.

To evaluate the effectiveness of JFD\textsuperscript{3},  we construct a new benchmark, named IRBlurUAV. It comprises 30,000 pairs of synthetically blurred and sharp IRUT images (IRBlurUAV-syn) and 4,118 real-world blurred images (IRBlurUAV-real), covering diverse motion directions, blur intensities, multi-scale UAV targets, and complex backgrounds.
Extensive experiments demonstrate that our JFD\textsuperscript{3} significantly outperforms state-of-the-art methods in detecting IRUT under motion blur.

Our main contributions are summarized as follows:
\begin{itemize}
    \item  We propose a joint framework, JFD\textsuperscript{3}, that unifies feature-domain restoration and IRUT detection in an end-to-end manner. The restoration component is optimized to recover features beneficial for detection, guided by detection objectives rather than generic visual quality. This task-driven design enhances detection performance under motion blur conditions. To the best of our knowledge, this is the first work to address IRUT detection  under motion blur conditions in a unified framework.
    \item We first introduce a feature-domain restoration strategy  specifically designed for infrared blurred images with a focus on the needs of target detection tasks. It focuses on restoring features relevant to detection. This method enhances the representation of target features in degraded images and improves detection performance.
    
    \item We design a novel frequency structure guidance module that integrates target frequency structural prior from the deblurring network into the detection backbone. This integration enhances the structural representation of IRUT under blur conditions by supplementing high-frequency structural details. The module significantly improves local discriminability and target localization.
\end{itemize}
\section{ Related work}
\subsection{Image Deblurring Methods}

Image deblurring has long been a core low-level vision task, essential for enhancing the quality of degraded visual inputs.
Early convolutional neural network(CNN)-based methods \cite{mao2023DeepRFT} directly map blurry images to sharp ones, forming the basis of many encoder–decoder architectures.
Transformer-based models \cite{chen2025MDT} enhance performance through long-range dependency modeling but introduce greater complexity.
Diffusion-based methods \cite{ren2023multiscale} and recent Mamba-style \cite{kong2025EVSSM,li2025MAIR} architectures have demonstrated impressive perceptual quality and generality, but remain computationally intensive.
However, most existing deblurring methods are rarely integrated with high-level tasks like detection. In contrast, our approach restores detection-relevant features in the feature domain to better support downstream task. 

\subsection{Infrared UAV Target Detection Methods}

In recent years, several IRUT detection methods \cite{rozantsev2016detecting,zhao2023infrared,fang2023danet} have been proposed, most of which focus on detection with clean images. And the majority of publicly available IRUT datasets \cite{huang2023anti,jiang2021anti,zhao2022vision} also consist of sharp, high-quality images.
To the best of our knowledge, UniCD \cite{fang2025detection} is the only work that explicitly considers degradation caused by temperature-dependent low-frequency nonuniformity for IRUT detection.
However, another common degradation in IRUT detection, motion blur, remains a largely unexplored gap in the field.
In this work, we aim to alleviate this gap by proposing the first end-to-end framework that jointly performs feature-domain deblurring and IRUT detection.

\begin{figure*}[!t]
\centering
\includegraphics[width=0.9\textwidth]{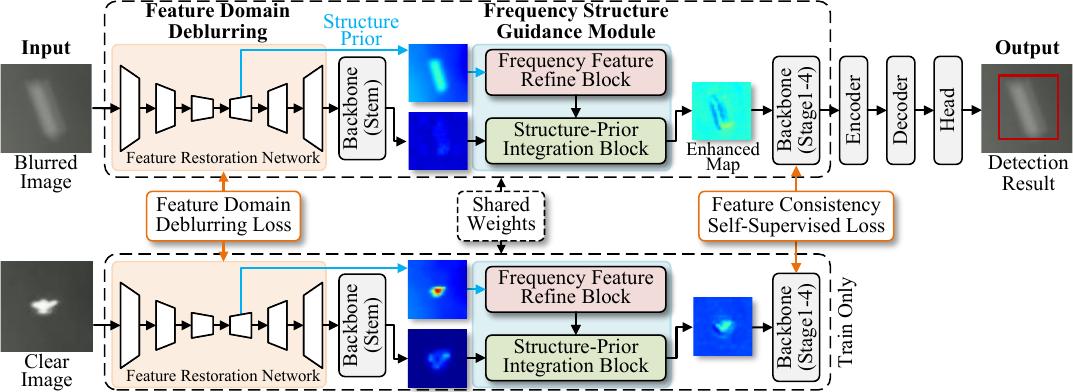} 
\caption{Overview of the proposed JFD\textsuperscript{3}, which first enhances degraded features through feature-domain restoration and then refines structural information using the frequency structure guidance module. The clear image branch supervises the blurred image branch using feature restoration loss and feature consistency self-supervised loss.}
\label{fig2}
\end{figure*}

\subsection{Joint Deblurring and Detection Methods}

Recent studies \cite{liu2022image,li2023detection} have begun to explore the integration of low-level restoration with high-level visual tasks, though most focus on haze and visible-light scenarios. 
Sayed et al. \cite{sayed2021improved} enhance motion-blurred detection using five classes of remedies.
Aakanksha et al. \cite{rajagopalan2023improving} introduce class-centric motion blur augmentation for segmentation.
DREB-Net \cite{li2025DREB} proposes a dual-stream fusion architecture for visible car targets detection under motion blur.
However, these methods overlook the unique challenges of infrared small-object detection under motion blur, such as texture scarcity and fine structural degradation.
To address this, we propose a frequency-guided module that enhances structural cues of small targets and improves robustness under blur degradation.


\section{The Proposed Method}

\subsection{Dual-Branch Joint Learning Framework}
\label{section:subsec1}

To address the challenge of degraded discriminative features under motion blur in IRUT images, we propose a dual-branch joint learning framework that enables feature-domain deblurring and detection to be collaboratively optimized in an end-to-end manner. The core design philosophy is to leverage clear-image supervision during training to guide the learning of robust representations, thereby enhancing the model's detection performance under blur degradation.

As illustrated in Figure \ref{fig2}, the proposed framework consists of two parallel branches: a clear-image branch and a blurred-image branch, which share weights to ensure feature space alignment. During training, both branches are activated. The clear branch operates on clear input and provides high-quality feature guidance, while the blurred branch is exposed to motion-degraded input and learns to restore and align its representations through supervision. During inference, only the blurred-image branch is retained.

Each branch begins with a lightweight feature restoration network, designed to compensate for low-level degradation. The restored feature maps are then passed through the detection network, consisting of a stem and multiple residual stages. We adopt DEIM \cite{huang2025deim} as our base detection architecture. Between the stem and the first stage of backbone, we integrate a frequency structure guidance module, which injects refined structural prior to enrich target localization cues. This design enables joint optimization of feature deblurring and detection tasks.

To enhance the network's feature extraction capability, we introduce a feature consistency self-supervised loss between the detection backbones of the blurred and clear branches. This loss encourages the blurred branch to produce intermediate representations that align with those from the clear branch, thereby improving its robustness under motion blur.

In contrast to other works that either directly detect from degraded input or treat deblurring as an isolated preprocessing step, our joint framework achieves task-aware feature enhancement through collaborative supervision and structure-guided information flow, ultimately yielding superior detection performance under blur conditions.

\subsection{Feature-Domain Deblurring (FDD) Network}

Conventional image-domain deblurring methods often incur significant computational cost and introduce redundant visual details irrelevant to detection. In contrast, we adopt a feature-domain deblurring strategy, which restores semantically meaningful representations directly in the latent space, thus enabling efficient and task-aligned enhancement.

As shown in Figure \ref{fig2}, the feature restoration network operates on the blurred image and refines it via a compact encoder–decoder structure. We build this module upon MIMO-UNet \cite{Cho_2021_MIMO}, a general-purpose deblurring network, and adapt it to our scenario by reducing the base channel number to 2 and the number of residual blocks per stage to 2. This design focuses on regulating the feature distribution, promoting semantic consistency with the clear branch, and reducing domain shift in the representation space—all while maintaining a low computational footprint suitable for real-time applications.

To guide the feature-domain restoration process, we design a two-part loss function that supervises both the encoder and decoder stages of the restoration network. Specifically, we enforce feature-level alignment between the blurred and clear branches using an $L_1$ loss in the encoder, and emphasize structural consistency in the decoder using a structural similarity-based loss.

Let $E_b^i$ and $E_c^i$ denote the intermediate feature maps from the $i$-th encoder stage of the blurred and clear branches, respectively. Similarly, let $D_b^j$ and $D_c^j$ denote the outputs from the $j$-th decoder stage. To preserve structural prior during decoding, we compute the structural similarity index measure (SSIM) between corresponding decoder features, which encourages the decoder to retain structural patterns rather than pixel similarity, thus promoting more robust feature representations for downstream detection. The final total feature-domain deblurring loss $\mathcal{L}_{deb}$ can be expressed by the following formula:
\begin{equation}
\mathcal{L}_{deb} = 
\sum_{i=1}^{3} \| E_b^i - E_c^i \|_1 +
\sum_{j=1}^{3} ( 1 - SSIM(D_b^j, D_c^j) ).
\end{equation}


\subsection{Frequency Structure Guidance Module }
\begin{figure}[t]
\centering
\includegraphics[width=0.9\columnwidth]{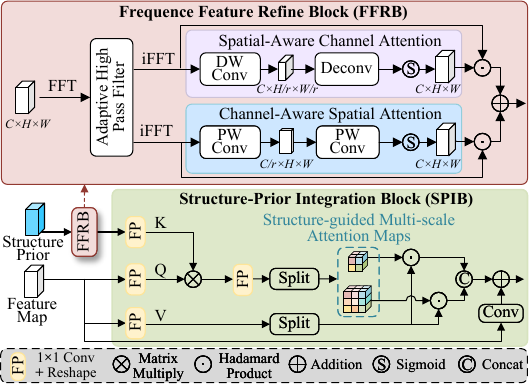} 
\caption{Overview of FSGM. The FFRB processes prior through high-pass filtering and attention mechanisms to refine feature representations. The SPIB integrates refined structure prior into the feature map.  }
\label{fig3}
\end{figure}

In blurred infrared UAV imagery, small target regions often suffer from degraded boundary structures and suppressed details. To address this, we incorporate a frequency structure guided module (FSGM) into our framework, which refines high-frequency structural prior and injects it into the detection backbone. Specifically,  we take the output feature map of the first decoder in the feature-domain deblurring network as the structural prior $P$. This feature map contains relatively rich structural and semantic information and is downsampled by a factor of 4 compared to the original image, which matches the resolution of the feature maps output by the stem and stage 1 of the detection backbone. The FSGM consists of two subcomponents, as shown in Figure \ref{fig3} : the frequency feature refine block (FFRB) and the structure prior integration block (SPIB).
Together, they extract, enhance, and integrate frequency-domain structure cues into feature representations for improved target discrimination.

\noindent\textbf{Frequency Feature Refine Block (FFRB).}
The FFRB aims to enhance discriminative details by extracting and refining high-frequency structural components $P_{\text{high}}$ from the structure prior $P$. We first transform $P$ into the frequency domain via the fast Fourier transform (FFT), then apply a learnable high-pass filter $\mathcal{H}_{\text{high}}(\cdot)$ to suppress low-frequency components. The filter is initialized with a frequency threshold of 0.5 and later adjusted adaptively during training. This filter retains the critical high-frequency details that are crucial for distinguishing small targets. Finally, the filtered result is transformed back to the spatial domain using the inverse fast Fourier transform (iFFT). The process is defined as:
\begin{equation}
P_{\text{high}} = \text{iFFT}\left(\mathcal{H}_{\text{high}}(\text{FFT}(P))\right),
\end{equation}

Next, we employ two types of attention mechanisms to refine the feature map: spatial-aware channel attention (SCA) and channel-aware spatial attention (CSA). These mechanisms are designed to enhance the feature map in a manner that prevents excessive compression in either the spatial or channel dimensions, especially for small target detection \cite{Dai_2021_ACL}.  Unlike traditional global attention that compresses spatial or channel dimensions into a single scalar, which may erase critical fine-grained cues, we adopt partial compression to retain target-relevant details, especially important for small objects. The refined prior $ P_{\text{refined}}$ is computed as:
\begin{equation}
 F_{\text{SCA}} = P_{\text{high}} \odot \sigma\left(  \mathrm{DeConv}(\mathrm{DWConv}(P_{\text{high}}))  \right),  
 \end{equation}
\begin{equation}
F_{\text{CSA}} = P_{\text{high}} \odot \sigma\left( \mathrm{PWConv}(\mathrm{PWConv}(P_{\text{high}})) \right),
\end{equation}
\begin{equation}
P_{\text{refined}}= F_{\text{SCA}} + F_{\text{CSA}},
\end{equation}
where $\mathrm{DWConv}$, $\mathrm{DeConv}$, and $\mathrm{PWConv}$ refer to depth-wise convolution, deconvolution, and point-wise convolution operations, respectively; \( \sigma(\cdot) \) is the sigmoid activation function; and \( \odot \) indicates element-wise multiplication.



\noindent\textbf{Structure Prior Integration Block (SPIB).}The SPIB takes as input the structure prior $ P_{\text{refined}}$ and the intermediate feature map $ f $. 
Both streams are first passed through feature projection (denoted as FP), consisting of a $1 \times 1$ convolution followed by reshaping. This yields the attention components: $ Q = \phi_q(f)$, $ K = \phi_k(P_{\text{refined}})$, and $ V = \phi_v(f)$, where $\phi$ denotes the projection operator.

The cross-attention matrix is then computed as $A = Q^\top K$. 
It produces pairwise correspondence between query and structure-guided keys. To inject multi-scale spatial cues, the attention map $A$ is split into two branches, denoted as $A_1$  and $A_2$  which function as structure-guided multi-scale attention maps corresponding to $5\times5$ and $ 7\times7$ dynamic kernels, respectively. In parallel, the projected value feature $ V$ is split into two matching branches $ V_1$  and $V_2$, ensuring alignment with the attention maps. Each attention map then modulates its corresponding value feature through element-wise Hadamard product, and the fused result is aggregated by channel-wise concatenation. Finally, a residual connection with $3\times3$ convolution from the original feature map $f$ is added to enhance gradient flow and representation fidelity:
\begin{equation}
F_{\text{PG}} =  \mathrm{Concat}(A_1 \odot V_1, A_2 \odot V_2) + \mathrm{Conv}(f).
\end{equation}
The resulting feature $F_{\text{PG}}$ is then forwarded to the subsequent detection network.
This design enables hierarchical integration of structure-aware guidance into the blurred feature representation.

\subsection{Joint Deblurring and Detection Loss} 

To enhance detection performance under motion blur, we adopt a multi-loss optimization strategy. Specifically, our framework is trained with the combination of three loss components: detection loss $\mathcal{L}_{\text{det}}$, feature deblurring loss $\mathcal{L}_{\text{deb}}$, and feature consistency self-supervised (FCSS) loss $\mathcal{L}_{\text{FCSS}}$. These components collaboratively supervise the dual-branch network, ensuring effective knowledge transfer from clear images to degraded ones, and facilitating the joint optimization of feature deblurring and detection.


The detection loss $\mathcal{L}_{\text{det}}$ is calculated based on the predictions from the blurred branch, following the original DEIM~\cite{huang2025deim} formulation. It serves as the task-specific objective to guide the end-to-end optimization.

To further bridge the representation gap across branches, we introduce the FCSS loss, which constrains the intermediate features of the blurred branch to align with their clear-image counterparts. For each stage $i$ in the shared detection backbone, we extract intermediate feature maps $F_C^{(i)}$ and $F_B^{(i)}$ from the clear and blurred branches, respectively. The consistency loss across all stages is then averaged as:
\begin{equation}
\mathcal{L}_{\text{FCSS}} = \frac{1}{4} \sum_{i=1}^{4} \left(1 - \frac{\mathbf{F}_C^{(i)} \cdot \mathbf{F}_B^{(i)}}{\|\mathbf{F}_C^{(i)}\| \|\mathbf{F}_B^{(i)}\|} \right).
\end{equation}

This self-supervised constraint promotes structural alignment and semantic consistency across branches. As supported by our network design, this encourages the blurred branch to better approximate clear-domain representations and facilitates more accurate detection under blur.

The overall training objective is defined as: $\mathcal{L}_{\text{total}} = \mathcal{L}_{\text{det}} + \lambda_1 \mathcal{L}_{\text{deb}} + \lambda_2 \mathcal{L}_{\mathrm{FCSS}}.$ Based on experience, we set the initial weights to $\lambda_1 = 0.4$, $\lambda_2 = 0.2$, and $\lambda_2 $ is annealed to 0.01 after 20 epochs, allowing the network to focus on detection accuracy in the following convergence phase.

\section{Experiments}

\subsection{Datasets and Evaluation Metrics}
\noindent\textbf{Datasets.} We construct a new benchmark dataset, IRBlurUAV, to facilitate the evaluation of IRUT detection under motion blur conditions. It comprises 30,000 pairs of synthetically blurred and sharp IRUT images (IRBlurUAV-syn) and 4,118 real-world blurred IRUT images (IRBlurUAV-real). All images have a size of 640×512. The synthetic images are generated using a process similar to that of \citet{sayed2021improved}, but with improved motion trajectory modeling that adopts linear paths with randomized directions and lengths to better approximate realistic UAV motion patterns. The dataset covers diverse backgrounds, multiple UAV scales, and various UAV types. All images are annotated with bounding boxes for detection tasks. The IRBlurUAV-syn set is split into training, validation, and test subsets using an 8:1:1 ratio, while IRBlurUAV-real serves exclusively as a test set  to evaluate generalization performance in real-world scenarios.  More details are provided in the supplementary material.

\noindent\textbf{Metrics.} We evaluate the model's detection performance using the standard COCO metrics: $AP$, $AR$, $AP_{50}$, and $AR_{50}$. $AP$ and $AP_{50}$ represent the detection precision over the IoU range of 0.50:0.95 and at IoU=0.50, respectively. $AR_{50}$ measures the average recall at IoU=0.50 with maxDets=100, and $AR$ represents the average recall over the IoU range of 0.50 to 0.95 with maxDets=1.  To assess model complexity, we consider the number of Parameters (Params), floating-point operations (FLOPs), and frames per second (FPS) for real-time performance. Additionally, we employ the Signal-to-Clutter Ratio (SCR) to evaluate the enhancement of the target signal in the feature domain.

\subsection{Experimental Details} The experiments are conducted on an NVIDIA RTX 3090 GPU with CUDA 12.1 and PyTorch 2.7. The model was trained for 150 epochs using the AdamW optimizer, with all other settings consistent with DEIM. Due to the lack of widely adopted infrared-specific deblurring methods, we utilize several general-purpose methods: DeepRFT (CNN-based) \cite{mao2023DeepRFT}, MDT (Transformer-based) \cite{chen2025MDT}, MaIR \cite{li2025MAIR} and EVSSM (Mamba-based) \cite{kong2025EVSSM}. For object detection, we apply CNN-based methods, including YOLO11-N, YOLO11-L \cite{yolo11_ultralytics}, MSHNet \cite{liu2024MSHNet}, and PConv (YOLOv8-N version) \cite{yang2025pinwheel}, as well as Transformer-based methods such as RT-DETR (ResNet18 version) \cite{zhao2024detrs}, D-FINE (N version) \cite{peng2024dfine}, and DEIM (D-FINE-N version)\cite{huang2025deim}. MSHNet and PConv are specifically designed for infrared small target detection. Finally, DREB-Net \cite{li2025DREB} is also used for comparison as a joint deblurring and detection method. For fairness, we retrained all methods on the IRBlurUAV-syn, and conducted evaluations on both IRBlurUAV-syn and IRBlurUAV-real to assess performance and generalizability.

\subsection{Quantitative Results}

\begin{table*}[!t]
    \centering
    \fontsize{9}{10}\selectfont 
    \setlength{\tabcolsep}{4pt}
    \begin{tabular}{c|ccc|ccccccc}
        \toprule
       \multirow{2}{*}{Strategy} & \multicolumn{3}{c|}{Module }  & \multicolumn{7}{c}{Metrics} \\ 
        \cline{2-4}  \cline{5-11}
        & Deblurring & Detection & Pub'Year  & $AP_{50}\uparrow$ & $AR_{50}\uparrow$ & $AP\uparrow$ &$AR\uparrow $& Params (M)$\downarrow$ &FLOPs (G)$\downarrow$ &FPS$\uparrow$\\ 
        
        \midrule
        \multirow{7}{*}{Direct} &\multirow{7}{*}{/}
        
         & YOLO11-N &2024 &0.510 & 0.530 & 0.213 & 0.258 & \underline{10.2} & \textbf{2.7} &\underline{69.2}   \\
         && YOLO11-L &2024 & 0.551 & 0.565 & 0.232 & 0.282 & 86.6 & 25.3 & 39.3 \\
        & & RT-DETR&CVPR'24 & 0.716 & \underline{0.811} & \underline{0.369} & \underline{0.400} &  19.0& 30.2 & 50.2  \\
             
        && D-FINE & ICLR’25 & \underline{0.722} & 0.795 & 0.347 & 0.382 & \textbf{3.5 }& 3.5 &  45.8\\
        && DEIM & CVPR’25 & 0.654 & 0.734 & 0.290 & 0.330 &\textbf{3.5 } & 3.5 & 45.8  \\
        && MSHNet & CVPR‘24 & 0.358 & 0.428 & 0.099 & 0.148 &15.5  & 38.2 & 53.1 \\
        && PConv & AAAI’25 & 0.581 & 0.592 & 0.227 & 0.278 &12.4  &\underline{2.9}& \textbf{83.7} \\
        \midrule
        \multirow{8}{*}{Separate} &
          \multirow{2}{*}{DeepRFT} &RT-DETR & \multirow{2}{*}{AAAI’23}&0.673 & 0.749&	0.284	&0.329&36.1&110.1  &9.6 \\
                                 &  &D-FINE & &0.660&0.743 &0.255 & 0.303& 20.6&  83.4& 9.7 \\
          \cline{2-4}  \cline{5-11}
          &\multirow{2}{*}{MDT} &RT-DETR & \multirow{2}{*}{CVPR’25}& 0.195& 0.297&0.062&0.082&31.6 &591.8   &1.5 \\
                                 &  &D-FINE & &0.181 &	0.268&	0.052&	0.069  &16.1 &565.1 & 1.5\\  
         \cline{2-4}  \cline{5-11}
         &\multirow{2}{*}{EVSSM} &RT-DETR & \multirow{2}{*}{CVPR’25}& 0.636	&0.713	&0.244&	0.285&  35.3  & 822.6&0.6 \\
                                 &  &D-FINE & &0.589	 &0.662	 &0.194	 & 0.234&  19.8  & 795.9&0.6\\  
        \cline{2-4}  \cline{5-11}
         &\multirow{2}{*}{MaIR} &RT-DETR & \multirow{2}{*}{CVPR’25}& 0.342	&0.471	 &0.118	 &0.150 & 39.7&582.4&0.1 \\
                                 & &D-FINE & &0.292	 &0.403	&0.084	  & 0.114 & 24.2&555.7 &0.1\\  
               
        \midrule
        \multirow{2}{*}{Joint} 
        &  \multicolumn{2}{c}{DREB-Net} & TGRS'25 & 0.710 & 0.754 & 0.300 & 0.357 & 34.6 & 684.2 &10.9 \\
        & \multicolumn{2}{c}{\textbf{Our JFD\textsuperscript{3}}}& - & \textbf{0.767} & \textbf{0.850} & \textbf{0.428} & \textbf{0.458} &\textbf{ 3.5} & 4.7 & 25.7 \\
        \bottomrule
    \end{tabular}
    \caption{Performance comparison of various methods on IRBlurUAV-syn dataset. \textbf{Bold} and \underline{underline} indicate the best and the second best results, respectively.}
    \label{tab:exp_uav}
\end{table*}

\begin{figure*}[!t]
\centering
\includegraphics[width=\textwidth]{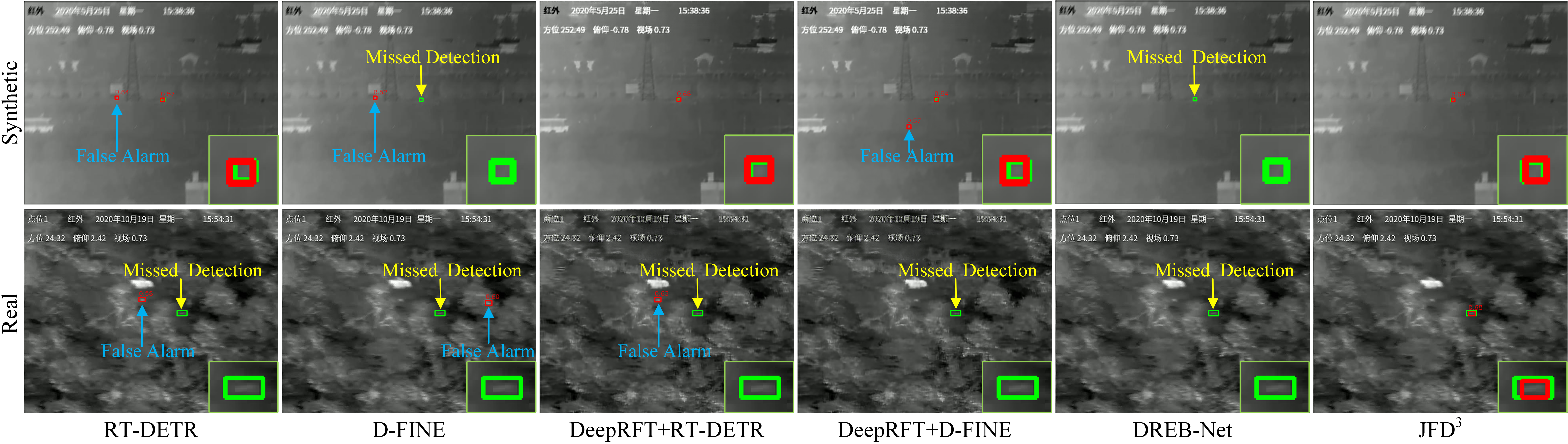} 
\caption{Comparison of detection results on IRBlurUAV-syn and IRBlurUAV-real, including direct, separate, and joint detection methods. Green and red boxes represent ground-truth and detected targets, respectively. Close-up views are shown in the bottom-right corner.}
\label{fig4}
\end{figure*}

\begin{table}[!t]
    \centering
    \fontsize{9}{9}\selectfont
    \setlength{\tabcolsep}{3pt}
    \begin{tabular}{c|cccc}
        \toprule
         Method & $AP_{50}\uparrow$ & $AR_{50}\uparrow$ & $AP\uparrow$ &$AR\uparrow $\\
        \midrule
        RT-DETR & 0.480 & 0.633 & 0.135 & 0.170  \\
        D-FINE & 0.514 & \underline{0.693} & \underline{0.151} & 0.190   \\
        DeepRFT + RT-DETR& 0.419 & 0.600 & 0.124 & 0.170  \\
        DeepRFT + D-FINE &0.437&0.638&0.129&0.181\\
        DREB-Net& \underline{0.520}&0.619&0.143 &\underline{0.196}\\
        \textbf{Our JFD\textsuperscript{3}} &\textbf{0.623} & \textbf{0.730}&\textbf{ 0.251} &\textbf{0.291}\\
        \bottomrule
    \end{tabular}
    \caption{Performance comparison of various methods on IRBlurUAV-real dataset.} 
    \label{tab:realblur}
\end{table}

As shown in Table \ref{tab:exp_uav}, when detecting blurred images directly, especially for infrared target detection methods like MSHNet and PConv, the loss of discriminative features between the target and the background due to motion blur is not considered, resulting in low accuracy and recall rates. When using separate detection methods, although the deblurring module restores the visual quality of the image, the deblurring process does not adequately focus on detection-friendly features. This leads to poor recovery performance in some methods, such as MDT, and even a significant decline in detection performance. Moreover, the deblurring modules in separate methods typically involve high computational complexity, limiting the overall real-time performance of the pipeline, thus reducing their efficiency.


In contrast to joint methods like DREB-Net which are not real-time, our approach achieves optimal detection performance at 25.7 FPS with high deployment efficiency, requiring only 120W power and 606 MB GPU memory on an RTX 3090. This is enabled by our lightweight modules (FDD and FSGM), which introduce only 0.02M parameters.

In Table \ref{tab:realblur}, we further validate the methods that performed well in Table 1 on the IRBlurUAV-real to test their performance under real-world blur conditions. The results show that our method still achieves the best detection performance when facing real blurred images, further demonstrating the robustness and practical value of our framework.

\subsection{Qualitative Results}

As shown in Figure \ref{fig4}, we present several effective methods, including direct detection methods (RT-DETR, D-FINE), methods that perform deblurring before detection (DeepRFT+RT-DETR, DeepRFT+D-FINE), as well as the joint method (DREB-Net) and our method. The first and second rows of the figure show the results of these methods on IRBlurUAV-syn and IRBlurUAV-real, respectively. Specifically, the third and fourth columns display deblurred images using DeepRFT, while the others show the blurred images.

From the results, it is evident that the target features are significantly degraded due to motion blur, making detection more difficult. Direct detection methods often lead to false alarms or missed detections because they struggle to distinguish discriminative features in blurred images. When deblurring is applied before detection, existing deblurring algorithms still struggle to recover degraded features under severe blur, further impacting detection performance. In contrast, our method can accurately detect UAV targets under motion blur conditions.

\subsection{Ablation Study}

This section presents ablation studies to validate the innovations of JFD\textsuperscript{3}. All experiments are conducted on the IRBlurUAV-syn. More experiments are provided in the supplementary material.

\begin{table}[t]
    \centering
    \fontsize{9}{9}\selectfont
    \setlength{\tabcolsep}{3pt}
    \begin{tabular}{cc|ccccc}
        \toprule
        FDD & FSGM& $AP_{50}\uparrow$ & $AR_{50}\uparrow$ & $AP\uparrow$ &$AR\uparrow $& SCR$\uparrow$   \\
        \midrule
        $\times$ & $\times$ & 0.654 & \underline{0.734} & 0.290 & 0.330 & 0.463 \\
        $\checkmark$ & $\times$ & \underline{0.763} & \textbf{0.852} & \underline{0.390} & \underline{0.426} & \underline{0.473}   \\
       $\checkmark$ & $\checkmark$ & \textbf{0.765} & \textbf{0.852} & \textbf{0.420} & \textbf{0.451} & \textbf{0.477}  \\
        \bottomrule
    \end{tabular}
    \caption{Ablation study of FDD and FSGM. }
    \label{tab:ablation_scr}
\end{table}

\noindent\textbf{Impact of FDD and FSGM.} As presented in Table 3, introducing the FDD module significantly improves all evaluation metrics, indicating its strong effectiveness in mitigating motion blur. This suggests that feature-domain deblurring plays a crucial role in enhancing detection under blurred conditions. When the FSGM is further added, the performance improves even more. This demonstrates that FSGM enhances the discriminability of target structures, complementing FDD and further boosting detection accuracy. Additionally, the SCR values across different stages of the backbone  reflect the improvement in target feature SCR, indicating enhanced feature extraction for detection. It can be seen that both of our proposed modules effectively enhance target features in the feature domain.

\begin{table}[!t]
\centering
\fontsize{9}{9}\selectfont
\setlength{\tabcolsep}{5pt}
\begin{tabular}{cc|cccc}
\toprule
IDD & FDD &   $AP_{50}\uparrow$ & $AR_{50}\uparrow$ & $AP\uparrow$ &$AR\uparrow $ \\
\midrule
$\times$ & $\times$ & 0.007 & 0.091 & 0.001 & 0.011  \\
\checkmark & $\times$ & 0.660& 0.743 & 0.255 & 0.303  \\
$\times$ & \checkmark & \textbf{0.763} & \textbf{0.852} & \textbf{0.390} & \textbf{0.426}  \\
\checkmark & \checkmark &\underline{0.703} & \underline{0.809} & \underline{0.307} & \underline{0.356}  \\
\bottomrule
\end{tabular}
\caption{Ablation study of IDD and FDD.}
\label{tab:ablation_deblurring}
\end{table}

\noindent\textbf{Impact of Image-Domain Deblurring (IDD) and Feature-Domain Deblurring (FDD).} Table  \ref{tab:ablation_deblurring} explores the relationship between image-domain deblurring and feature-domain deblurring, tested on IRBlur-syn. The first row represents our baseline architecture, trained on clear images without FDD and FSGM, and tested directly on blurred images to simulate real-world scenarios. The performance is almost zero, highlighting the significant impact of motion blur on detection that only considers clear images. The second row shows the results of applying DeepRFT to the blurred images before detection, which leads to some improvements in performance with the separate deblurring approach. The third row presents our final JFD\textsuperscript{3}, where feature-domain deblurring reaches optimal performance. The last row demonstrates the use of deblurred images as input for the JFD\textsuperscript{3}, showing a comprehensive improvement over the second row, indicating that feature-domain and image-domain deblurring complement each other. This combination enhances target features that image-domain deblurring alone could not recover effectively.

\begin{table}[!t]
\centering
\fontsize{9}{9}\selectfont
\setlength{\tabcolsep}{4pt}
\begin{tabular}{c|c|ll}
\toprule
\makecell{Blur Level\\ $\in$(PSNR range)}  & Method & \multicolumn{1}{c}{$AP_{50}\uparrow$ } & \multicolumn{1}{c}{$AR_{50}\uparrow$ } \\ 
\midrule

  \multirow{2}{*}{\makecell{Severe \\   $\in[10, 20)$}} 
  & DeepRFT  + RT-DETR & 0.542 & 0.638 \\ 
  & JFD\textsuperscript{3} & \textbf{0.562 }& \textbf{0.723} \\
\midrule
\multirow{2}{*}{\makecell{Moderate\\ $\in[20, 22.5)$}} 
  & DeepRFT + RT-DETR & 0.664 & 0.745 \\  
  & JFD\textsuperscript{3} & \textbf{0.672} & \textbf{0.788} \\
\midrule
\multirow{2}{*}{\makecell{Mild \\ $ \in[22.5, 32)$}}
  & DeepRFT + RT-DETR &\textbf{ 0.772} & 0.831 \\   
  & JFD\textsuperscript{3} & 0.767 & \textbf{0.853} \\
  
\bottomrule
\end{tabular}
\caption{Performance comparisons with different blur levels.}
\label{tab:blur_level_compare}
\end{table}

\noindent\textbf{Impact of Different Blur Levels.} In Table \ref{tab:blur_level_compare}, we categorize test sets into three levels of blur severity. For each blur level, we investigate the performance of both separate methods and our JFD\textsuperscript{3}. As the blur severity increases, the detection performance of all methods decreases, highlighting the growing interference caused by more severe blur. However, when examining each blur level individually, JFD\textsuperscript{3} demonstrates a greater performance improvement with increasing blur severity. Specifically, for more severe blur, JFD\textsuperscript{3} outperforms the separate methods, indicating that our approach is more effective in handling severe blur conditions.

 \begin{table}[!t]
    \centering
    \fontsize{9}{9}\selectfont
    \setlength{\tabcolsep}{4pt}
    \begin{tabular}{c|cc|cc}
        \toprule
         \multirow{2}{*}{Module} & \multicolumn{2}{c|}{w/o FDD} &\multicolumn{2}{c}{w/ FDD}\\
           & PSNR & SSIM  & PSNR  & SSIM \\
        \midrule
         Stem         & 14.66 & 0.5977 & \textbf{15.64}$\uparrow$ & \textbf{0.6201}$\uparrow$\\
         Stage1    & 15.29 & 0.5189  & \textbf{18.18}$\uparrow$ & \textbf{0.6464}$\uparrow$\\
        \bottomrule
    \end{tabular}
    \caption{Impact of FDD  at different backbone layers.}
    \label{tab:feature_injection_ablation}
\end{table}

\noindent\textbf{Impact of FDD at Backbone Shallow Layers.} As shown in Table \ref{tab:feature_injection_ablation}, we evaluate the effect of FDD by calculating the Peak Signal-to-Noise Ratio (PSNR) and Structural Similarity Index Measure (SSIM) between the shallow-layer feature maps extracted from clear and blurred inputs, with and without FDD. The results show that FDD significantly improves both metrics, indicating enhanced discriminative quality of early-stage features and better overall representation.

More visualization results on IRBlurUAV and experimental results on other datasets can be found in the supplementary material.

\section{Conclusion}
In this paper, we propose the JFD\textsuperscript{3}, an end-to-end dual-branch framework for IRUT detection under motion blur. First, we introduce a lightweight feature restoration network that focuses on feature-domain deblurring. Next, we propose a frequency structure guidance module that enhances target structural information beneficial for detection. 
Additionally, we construct a dataset named IRBlurUAV with diverse motion blur infrared UAV images. Experiments show that JFD\textsuperscript{3} outperforms existing approaches in both simulated and real-world scenarios, while maintaining real-time performance.

\section{Acknowledgments}
This work was supported by the Open Research Fund of the National Key Laboratory of Multispectral Information Intelligent Processing Technology under Grant 61421132301, the Natural Science Foundation of Jiangsu Province BK20232028, and in part by the projects of the National Natural Science Foundation of China under Grants No. 62472341, 62372351, U21B2015, 62371203 and 62301228.




\bibliography{aaai2026}

\end{document}